%% file: main.tex
\documentclass[10pt,twocolumn,letterpaper]{article}

\usepackage{iccv}              % To produce the CAMERA-READY version

\definecolor{iccvblue}{rgb}{0.21,0.49,0.74}
\usepackage[pagebackref,breaklinks,colorlinks,allcolors=iccvblue]{hyperref}
\usepackage{array} % required for text wrapping in tables

\usepackage[accsupp]{axessibility}  % Improves PDF readability for those with disabilities.

\def\paperID{*****} % *** Enter the Paper ID here
\def\confName{ICCV}
\def\confYear{2025}

\title{
AtomDiffuser: Time-Aware Degradation Modeling for Drift and Beam Damage in STEM Imaging
}

\author{Hao Wang$^1$, Hongkui Zheng$^2$, Kai He$^2$, and Abolfazl Razi$^1$\thanks{Corresponding author}\\
$^1$Clemson University\\
$^2$University of California, Irvine\\
{\tt\small \{hao9, arazi\}@clemson.edu, \{kai.he, zhongkui\}@uci.edu}
}
\begin{document}
\maketitle
\begin{abstract}
% Understanding atomic motion and beam-induced damage is critical for analyzing dynamic material transformations in scanning transmission electron microscopy (STEM). However, robust estimation of atomic drift and decay between consecutive frames remains a challenging task due to complex degradation processes and imaging noise. In this work, we introduce \textbf{AtomDiffuser}, a learning-based framework for joint prediction of atom drift and beam-induced damage using a synthetic degradation model. Given a reference STEM frame and a target frame at arbitrary time step, our model estimates the underlying affine motion field and pixel-wise decay map, effectively disentangling geometric transformations from signal loss. We simulate the degradation process with time-dependent affine perturbations, random atom removals, and noise injection to train the model in a supervised manner. Extensive expeiments on synthetic and real STEM sequences demonstrate that AtomDiffuser accurately recovers atom trajectories and localized damage patterns, enabling fine-grained analysis of material degradation. Our framework provides a novel foundation for predictive modeling of temporal evolution in atomically-resolved microscopy.

Scanning transmission electron microscopy (STEM) plays a critical role in modern materials science, enabling direct imaging of atomic structures and their evolution under external interferences. However, interpreting time-resolved STEM data remains challenging due to two entangled degradation effects: spatial drift caused by mechanical and thermal instabilities, and beam-induced signal loss resulting from radiation damage. These factors distort both geometry and intensity in complex, temporally correlated ways, making it difficult for existing methods to explicitly separate their effects or model material dynamics at atomic resolution.
In this work, we present AtomDiffuser, a time-aware degradation modeling framework that disentangles sample drift and radiometric attenuation by predicting an affine transformation and a spatially varying decay map between any two STEM frames. Unlike traditional denoising or registration pipelines, our method leverages degradation as a physically heuristic, temporally conditioned process, enabling interpretable structural evolutions across time. Trained on synthetic degradation processes, AtomDiffuser also generalizes well to real-world cryo-STEM data. It further supports high-resolution degradation inference and drift alignment, offering tools for visualizing and quantifying degradation patterns that correlate with radiation-induced atomic instabilities.
Our source code and sample datasets are available at: \url{https://arazi2.github.io/aisends.github.io/project/Atom}

\vspace{-0.5cm}
% \url{}
\end{abstract}

% Diffusion:
% Diffusion-based deep learning method for augmenting ultrastructural imaging and volume electron microscopy
% Denoising diffusion probabilistic models for generation of realistic fully-annotated microscopy image datasets

% Denoising:
% Physics Informed Neural Network Enhanced Denoising for Atomic Resolution STEM Imaging

% Precision controlled atomic resolution scanning transmission electron microscopy using spiral scan pathways
% Determination of atomic positions from time resolved high resolution transmission electron microscopy images
% Dynamic hetero-metallic bondings visualized by sequential atom imaging

% Revolving scanning transmission electron microscopy: Correcting sample drift distortion without prior knowledge
% Identification and Correction of Temporal and Spatial Distortions in Scanning Transmission Electron Microscopy
% Atomap: a new software tool for the automated analysis of atomic resolution images using two-dimensional Gaussian fitting
% Learning and Controlling Silicon Dopant Transitions in Graphene using Scanning Transmission Electron Microscopy
% Deep convolutional neural networks to restore single-shot electron microscopy images

\section{Introduction}
\label{sec:intro}

% \hw{What is STEM imaging, why matter}
Understanding how materials evolve at the atomic scale is central to modern materials science, with implications ranging from battery durability to semiconductor reliability \cite{zheng2022situ}. 
Many key properties, such as conductivity, mechanical strength, and chemical stability, are governed by atomic arrangements and their transformations over time \cite{xiao2024automated,sadri2024unsupervised}. 
% To observe these processes directly, researchers rely on advanced electron microscopy techniques that offer sub-nanometer spatial resolution and, increasingly, temporal resolution as well. 
Scanning transmission electron microscopy (STEM), transmission electron microscopy (TEM), and cryogenic variants (cryo-STEM/TEM) enable visualization of individual atomic columns and structural changes under controlled environments \cite{zheng2022situ,masud2024machine}. 
While these imaging modalities enable direct observation of atomic-scale degradation, their application to beam-sensitive materials remains challenging, as even low-dose electron exposure can lead to atomic displacement, defect formation, or irreversible chemical changes \cite{ghosh2022bridging,liao2024fast,weile2025defect}.
For instance, the garnet-type solid electrolyte $\mathbf{Li_7La_3Zr_2O_{12}}$ (LLZO) is a promising candidate for solid-state lithium batteries \cite{zheng2024convolutional}. However, its crystal lattice is prone to beam-induced degradation even under cryogenic temperatures, and tools for quantitatively analyzing such degradation pathways remain limited due to its current way of imaging.

\begin{figure}[htbp]
    \centering
    \includegraphics[width=1\linewidth]{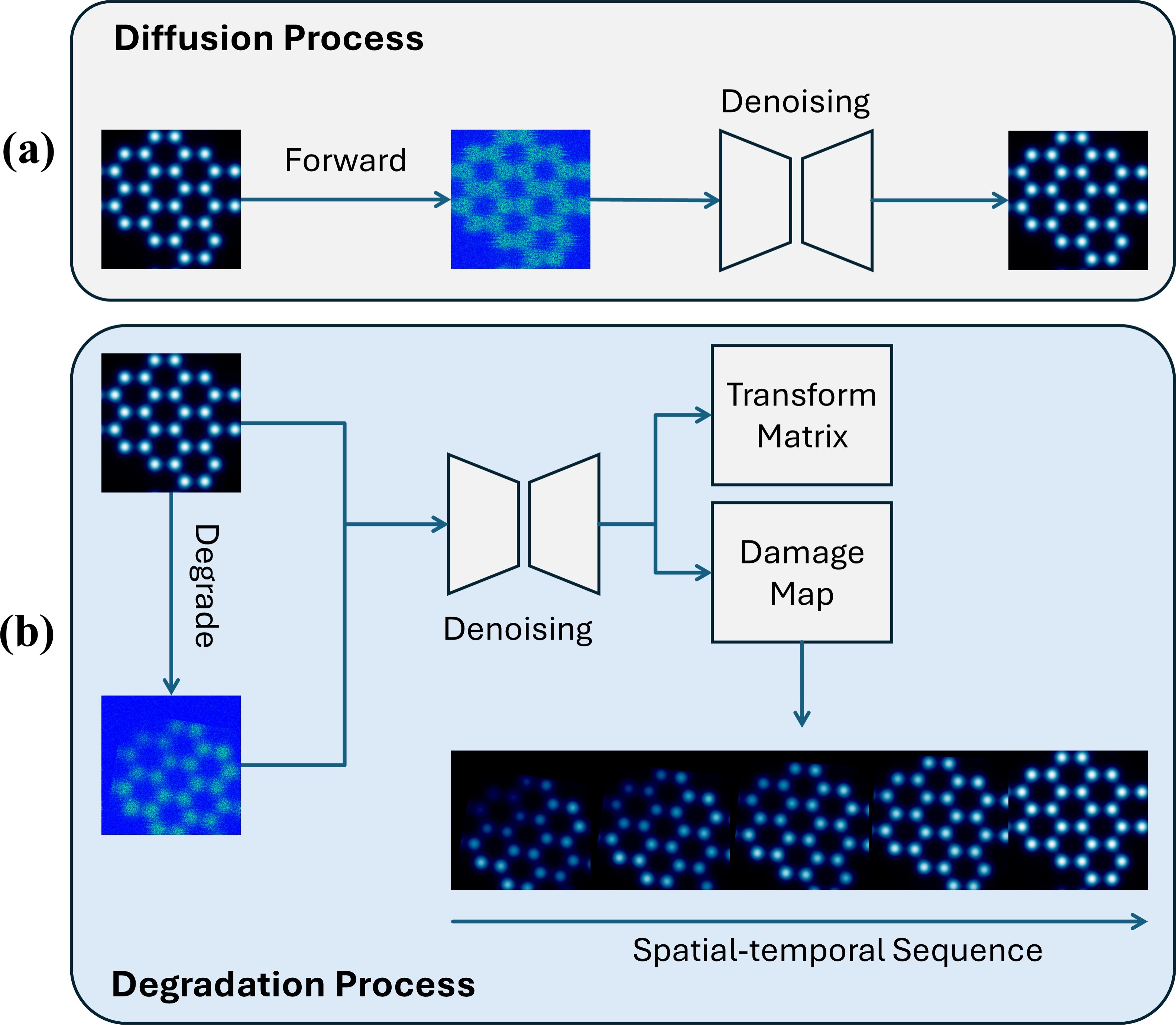}
    \caption{Differences between the diffusion process \cite{rombach2022high} and the degradation process \cite{bansal2023cold}.}
    \label{fig:cover}
    \vspace{+0.25cm}
\end{figure}

% \hw{What is the issue, how does it happen}
During atomic-resolution STEM imaging, two degradation mechanisms: atomic drift and beam-induced damage, intertwined, pose fundamental challenges to accurate structural interpretation \cite{kalinin2021automated}. 
Atomic drift arises from specimen charging, thermal expansion, or mechanical instabilities in the stage, leading to subpixel displacements of atomic columns across sequential frames. 
Although drift does not alter signal intensity, it distorts geometric structure and may mimic dynamic phenomena~\cite{kalinin2021automated}. 
In contrast, beam-induced damage stems from inelastic or elastic energy transfer, causing irreversible structural alterations such as atom ejection, radiolytic bond scission, or elemental loss~\cite{egerton2004radiation}. 
These effects manifest as intensity reduction, blurring, or structural disappearance. Since drift and damage often co-occur and evolve gradually during imaging, their entangled manifestations complicate the analysis of dynamic processes, particularly in beam-sensitive materials. 
Disentangling these two mechanisms is essential for reliable interpretation of material behavior under electron irradiation~\cite{ophus2016correcting}.

% \hw{How it was addressed, still work now?}
As beam-induced damage is readily quantifiable in the absence of motion, correcting for drift becomes the primary challenge in enabling such direct assessment.
Traditionally, the field has addressed drift using either hardware-level solutions, such as revolving scan paths\cite{sang2014revolving} or spiral scanning \cite{sang2017precision}, or post-acquisition correction algorithms based on scan orientation \cite{ophus2016correcting}.
% , lattice matching\cite{hussaini2018determination}, or principal component decomposition\cite{roccapriore2021identification}. 
These methods improve alignment and image quality but do not provide interpretable or predictive models of atomic motion itself.
More recently, automated tracking systems\cite{hussaini2018determination} and deep learning–based image restoration models\cite{ihara2022deep,lobato2024deep} have been applied to high-resolution STEM data, achieving impressive results in noise reduction and scan artifact correction. 
However, the drift effects addressed in these approaches are more akin to scan artifacts or optical distortions within a single frame, rather than true atomic motion across frames.
% However, these approaches typically treat drift as an implicit artifact to be suppressed, rather than as an explicit, structured transformation to be modeled.

% \hw{Our task, why it is different, why it is hard}
Unlike scan distortion or minor drift commonly addressed in single-frame correction methods, the geometric evolution in our STEM sequences reflects sample-driven atomic motion accumulated over long exposures \cite{zheng2024convolutional}. These sequences are temporally sparse, with frames acquired $1.6–2.0$ seconds apart, which is sufficient for beam-induced forces to cause substantial and spatially nonuniform displacements, often exceeding tens or hundreds of pixels. Moreover, because damage and drift co-occur, image features may vanish, deform, or shift in ways that are physically irreversible. This poses a fundamental modeling challenge: there is no stable reference frame, no consistent pixel-to-pixel correspondence, and no purely geometric transformation that can recover the original signal.

% \hw{AtomDiffuser}
To address these challenges, we present \textit{AtomDiffuser}, a diffusion-variant spatiotemporal degradation model that disentangles atomic drift and beam-induced damage from sequential STEM image pairs. 
As shown in Figure \ref{fig:cover}, given a reference frame, a target frame, and a time-step signal, the model predicts a global affine transformation for drift and an atomic-level attenuation map for beam damage. 
AtomDiffuser is trained on synthetic sequences that simulate time-dependent physical degradation, enabling it to learn the spatiotemporal dynamics of atomic evolution via a time-conditioned U-Net.
In evaluations of damage assessment, AtomDiffuser significantly outperforms the existing CNN method \cite{zheng2024convolutional}.
Applied to synthetic and cryo-STEM data of LLZO, it enables estimation of atomic drift and prediction of spatially resolved, interpretable beam-induced signal degradation.

Our main contributions are as follows:
\begin{itemize}
\item We introduce a time-conditioned lightweight deep learning model based on U-Net, together with a synthetic data generator that simulates realistic drift, stochastic decay, and acquisition noise during STEM imaging, reflecting key sources of temporal degradation that limit the interpretability of atomic-resolution experiments on beam-sensitive materials.
\item We propose a degradation learning framework that formulates inter-frame atomic motion as a global affine transformation, enabling explicit and interpretable drift estimation across arbitrary time intervals, beyond conventional flow or registration methods, which is critical for compensating specimen instabilities in long-exposure or time-resolved microscopy studies.
\item We model beam-induced degradation as a differentiable, pixel-wise attenuation process, allowing fine-grained prediction of atomic damage aligned with physical decay mechanisms, providing a step toward spatially resolved damage mapping that could inform dose management and acquisition strategies in in-situ STEM experiments.
\end{itemize}

% Our main contributions are as follows:
% \begin{itemize}
%     \item We introduce a time-conditioned lightweight deep learning model that is based on U-Net, and a synthetic data generator that simulates realistic drift, stochastic decay, and acquisition noise during STEM imaging.
%     \item We propose a degradation learning framework that formulates inter-frame atomic motion as a global affine transformation, enabling explicit and interpretable drift estimation across arbitrary time intervals, beyond conventional flow or registration methods.
%     \item We model beam-induced degradation as a differentiable, pixel-wise attenuation process, allowing fine-grained prediction of atomic damage aligned with physical decay mechanisms.
%     % \item Our model supports drift and damage interpolation at arbitrary exposure steps, enabling temporal super-resolution and aiding downstream analysis of material stability.
% \end{itemize}

\section{Related Work}

\subsection{Atomic Feature Extraction from STEM Images}
Early attempts at atom localization were severely affected by drift distortion, which was addressed through RevSTEM \cite{sang2014revolving}, a method that estimates and corrects sample drift using rotational scan series without requiring prior structural knowledge.
Subsequent classical pipelines incorporated a model-free correction method based on orthogonal scan pairs to address nonlinear scan distortions\cite{ophus2016correcting}.
Meanwhile, a specialized scan strategy is applied to reduce distortion, such as the spiral scanning approach, which mitigates fly-back and nonlinear artifacts by replacing raster trajectories with smooth, continuous spiral paths \cite{sang2017precision}.
To enable high-precision quantification of atomic fluctuations from time-resolved TEM videos, an automated image processing pipeline was developed that combines template matching, subpixel centroid localization, and geometric modeling to track atom positions and structural evolution in catalyst nanoparticles \cite{hussaini2018determination}.
With the advent of deep learning, convolutional networks have become the de facto tools for predicting pixel-wise atomic column probability maps under low-dose conditions~\cite{lin2021temimagenet}. 
These models substantially improve the recall of weak-contrast atoms and support downstream tasks such as lattice reconstruction, defect counting, and strain mapping \cite{dey2024addressing}.  
Nevertheless, most existing networks assume static scenes. When consecutive frames are compared, the entangled effects of beam-induced attenuation and specimen drift are typically addressed through post-hoc intensity differencing. As a result, they cannot produce interpretable, parameterized estimates of the two degradation mechanisms that drive structural evolution.

Recent efforts have also addressed atomic-level distortion using statistical modeling techniques, such as Gaussian modeling\cite{zheng2025gaussian} and Gaussian process regression~\cite{roccapriore2021identification}. Deep learning has been applied to suppress scan-induced artifacts~\cite{ihara2022deep,lobato2024deep}, reconstruct atomic structures from noisy observations~\cite{he2021hybrid,lee2025using}, and track atom centers with high precision. 
Industrial-ready toolkits such as AtomAI and AtomSegNet have further enabled automated atomic structure reconstruction and large-scale analysis of STEM data~\cite{ziatdinov2022atomai,lin2021temimagenet}.
Beyond reconstruction, deep models have been used to simulate complex physical processes, including graphene healing dynamics~\cite{ghosh2022bridging}, defect detection in van der Waals materials~\cite{weile2025defect}, and recovery of molecular 3D motion from electron microscopy videos~\cite{ye2023recovering}. 
To the best of our knowledge, no prior work explicitly models atomic drift and attenuation as coupled, learnable degradation processes across frames.

\subsection{Diffusion Modeling}
Denoising diffusion probabilistic models (DDPMs) have recently emerged as the dominant paradigm for high-quality image synthesis across a wide range of computer vision tasks \cite{rombach2022high}. Compared to classical Generative Adversarial Networks (GANs), diffusion models offer greater stability and flexibility, particularly in conditional generation settings where guidance signals such as text prompts or semantic maps can be incorporated~\cite{choi2023custom,mokady2023null}. These capabilities have enabled applications in digital art~\cite{zhang2025soda}, content creation~\cite{meng2021sdedit,zhou2023denoising}, and industrial data augmentation~\cite{Wang_2025_WACV}, where semantic control and visual fidelity are critical.

In the materials science domain, the EMDiffuse family \cite{lu2023emdiffuse} first demonstrated that fine-tuning a generic DDPM on as little as a single noisy–clean image pair can outperform CNN-based denoisers on cryo-EM and cryo-STEM data, while additionally providing per-pixel uncertainty estimates. Its super-resolution variant, EMDiffuse-r \cite{lu2024diffusion}, synthesizes high-resolution 3D volumes from anisotropic stacks, enabling isotropic reconstruction without hardware modification.
More recently, SARDiffuse \cite{qiu2025deep} introduced a diffusion-based approach for correcting lens-aberration blur in uncorrected STEM images, achieving routine sub-\AA{}ngström resolution and accurate atomic localization. Other diffusion models have been applied to reconstruct crystal structures from noisy data \cite{pakornchote2024diffusion} and to simulate irradiation-induced defect cascades in solids \cite{liao2024fast}, further highlighting the versatility of probabilistic diffusion frameworks for materials research.

Despite the superior performance of DDPM in physics modeling of atomic images, current diffusion models for EM/STEM heavily focus on correcting atom distortion rather than inter-frame drift \cite{belardi2025improving}, and prefer to analyze damage only on single images instead of temporally. Works that can observe material evolution over time while explicitly disentangle geometric drift from radiometric attenuation are still rare. 

 % and score-based generative models reconstruct data by inverting a learned noise-adding process \cite{}.  Compared with GANs, they train stably and reproduce high-frequency detail without mode collapse, which is critical for atomic-scale imaging where every pixel carries structural information.
% Diffusion priors have likewise been exploited to inpaint sparsely sampled line-hop scans, markedly reducing cumulative beam dose while preserving atomic detail \cite{nicholls2020minimising}.
% Complementary to generative restoration, \cite{moshtaghpour2025diffusion} model beam damage itself as a diffusion-style partial differential equation, deriving a Diffusion Distribution Model that predicts how radiolytic or ballistic damage halos overlap during a raster scan.  Their analysis motivates diffusion-aware scan patterns that lower peak damage by spreading probe dwell locations.
% Our work bridges this gap: we view the transition between two STEM frames as a \emph{time-conditioned degradation process} governed by an affine drift field and a spatially varying attenuation map, both regressed by a dual-headed diffusion-inspired network.  This enables, for the first time, joint, interpretable estimation of atom drift and beam damage from a single image pair.

\section{Methodology}
\begin{figure}[htbp]
\vspace{-0.25cm}
    \centering
    \includegraphics[width=1\linewidth]{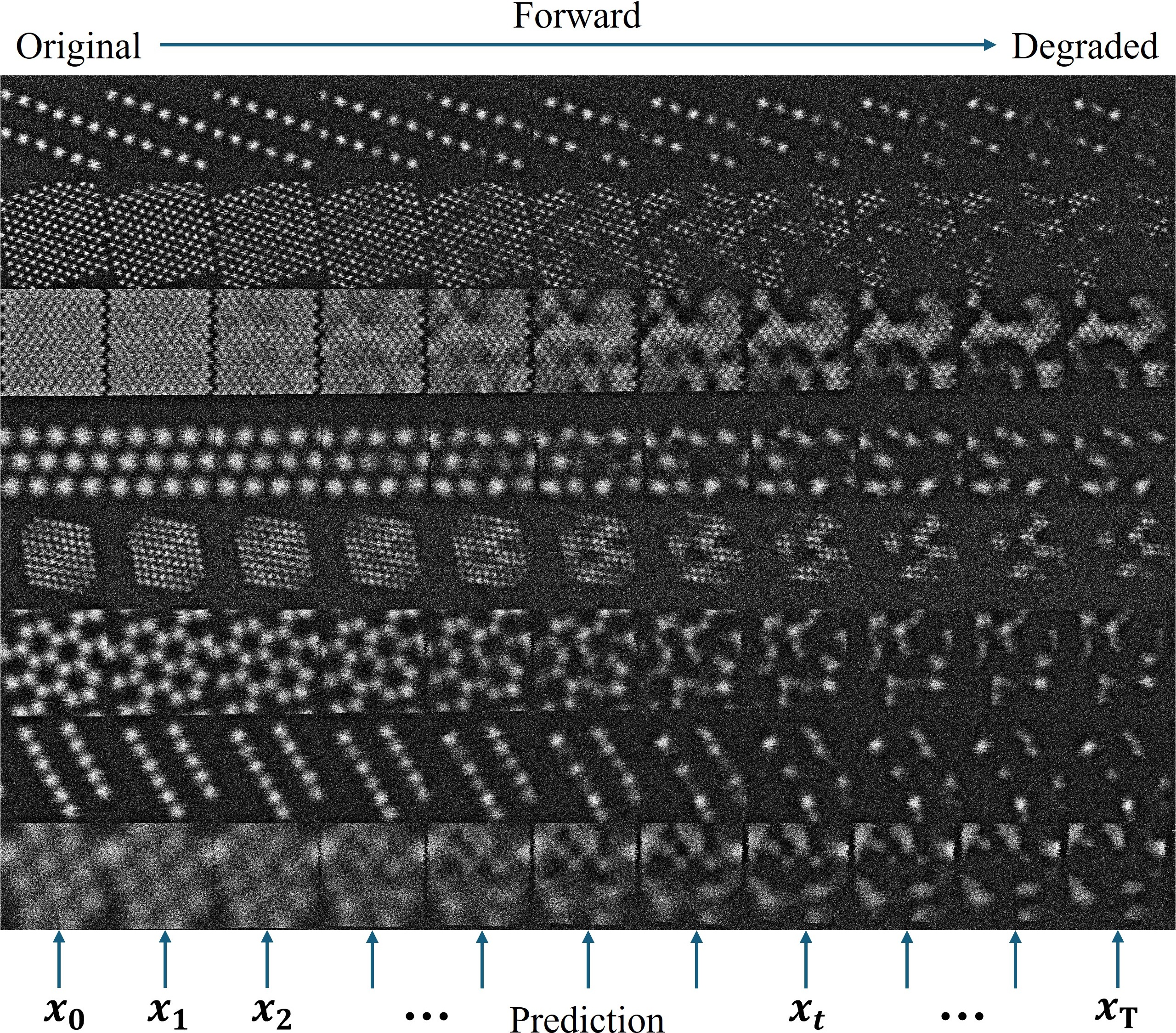}
    \caption{Degradation modeling. From left to right, the STEM samples are degraded over time.}
    \label{fig:degrade}
    \vspace{-0.5cm}
\end{figure}

\begin{figure*}[htbp]
    \centering
    \includegraphics[width=1\linewidth]{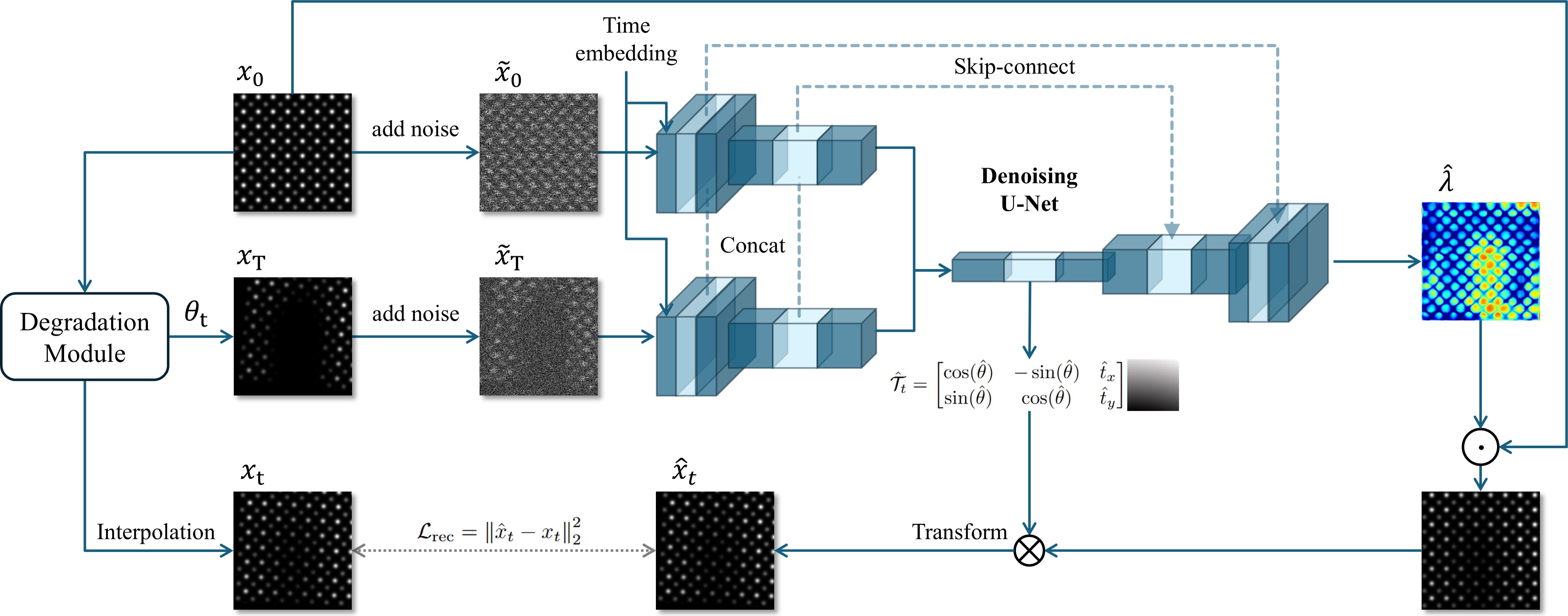}
    \caption{Framework of the proposed AtomDiffuser.}
    \label{fig:frame}
    \vspace{-0.5cm}
\end{figure*}

\subsection{Degradation Modeling}
End-to-end prediction of atomic changes in STEM sequences, such as directly regressing $x_t$ from $x_0$, is often unreliable due to the presence of noise, sparsity, and large-scale rigid motion \cite{zheng2022situ}. 
% These challenges are further amplified by the irreversible nature of beam-induced degradation, where structural information is progressively lost over time. 
To address this, we adopt a degradation modeling approach inspired by diffusion models, enabling continuous and physically interpretable modeling of structural evolution.

While classical diffusion models have shown strong performance in generative and denoising tasks by learning to reverse a stochastic corruption process \cite{lu2024diffusion}, they are less suited for modeling physically driven phenomena such as drift or beam damage. 
In DDPM, for example, the forward diffusion is defined by progressively adding Gaussian noise to $x_0$:
\begin{align}
x_t = \sqrt{\bar{\alpha}_t} x_0 + \sqrt{1 - \bar{\alpha}_t} \, \epsilon, \quad \epsilon \sim \mathcal{N}(0, \mathbf{I}),
\end{align}
where $\bar{\alpha}_t$ denotes the cumulative product of the noise schedule, and the noise factor $\epsilon$ is sampled from a standard Gaussian distribution $\mathcal{N}$. 
However, this formulation is limited in its ability to incorporate interpretable, structured degradation signals such as rigid motion, as DDPMs are primarily designed for noise removal rather than physical factor estimation.

In contrast, deterministic degradation models, such as Cold Diffusion \cite{bansal2023cold}, replace the stochastic noise process with a physically interpretable degradation process:
\begin{align}
x_t = \mathcal{D}(x_0, t),
\end{align}
where $\mathcal{D}$ may represent operations such as blur, masking, or spatial warping, as shown in Figure \ref{fig:degrade}. 
This formulation preserves temporal consistency and structural semantics while aligning more naturally with real-world degradation patterns, allowing the degraded state $x_t$ to be computed directly from any time step $t$, which is consistent with the classical DDPM.

Inspired by this degradation modeling, we define a physically structured degradation function that models the evolution of STEM images over time as a combination of atomic drift and signal attenuation:
\begin{align}
x_t  = \mathcal{D}(x_0, t) = \Phi(\lambda_t \odot  x_0;\ \mathcal{T}_{t}), 
\end{align}
where $\Phi$ applies a spatial transformation ($\mathcal{T}_t \in \mathbb{R}^{2 \times 3}$) to account for atomic drift that maps from $x_0$ to $x_t$, and $\lambda_t \in [0,1]^{H \times W}$ denote a pixel-wise attenuation map to model beam-induced signal loss, where each value can be interpreted as the relative survival coefficient of the corresponding atomic column over time under cumulative electron exposure. 
% Given the clean reference $x_0$, our goal is to generate the degraded state $x_t$, same as the Cold-Diffusion, as shown in Figure \ref{fig:cover}.
Given a clean reference image \( x_0 \), our goal is to generate the degraded state \( x_t \), following the theory from the Cold-Diffusion, as illustrated in Figure~\ref{fig:cover}.

\subsection{Restoration}
Unlike Cold-Diffusion models, our objective is not to reconstruct the clean image $x_0$, but rather to recover the latent physical factors, specifically, atomic drift and signal decay, that cause $x_0$ to degrade into $x_t$:
\begin{align}
\hat{\lambda}_t, (\hat{\theta}, \hat{t}_x, \hat{t}_y)  &= \mathcal{f}(x_0, t; \ x_T),
\end{align}
where $(\hat{\theta}, \hat{t}_x, \hat{t}_y)$ denote the predicted parameters of in-plane rotation and translation, and $\lambda \in [0,1]^{H \times W}$ denote the predicted decay map. 

This fundamental distinction motivates a different conditioning strategy: we use $x_T$ as the structural prompt, and supervise the prediction of intermediate degradation states $x_t$ by learning to infer the parameters
% $(\lambda_t, \mathcal{T}_t)$ 
that explain the evolution from $x_0$ to $x_t$.

% To enable supervised learning under this degradation inversion framework, we construct a pipeline that simulates the drift physics and beam-induced degradation. 
Given the intermediate parameters $(\hat{\lambda}_t, \hat{\mathcal{T}}_t)$, we are able to reconstruct the degraded state $x_t$ by applying two successive transformations: 
\begin{align}
\label{eq:pred}
x_t' &= \hat{\lambda}_t \odot  x_0 \\
\hat{x}_t &= \Phi(x'_t;\ \hat{\mathcal{T}}_{t}),
\end{align}
where first, a spatially varying attenuation map $\hat{\lambda}_t$ to model signal decay, and second, an affine transformation $\hat{\mathcal{T}}_{t} \in \mathbb{R}^{2 \times 3}$ to model global drift, where
\begin{align}
\hat{\mathcal{T}}_{t}= 
\begin{bmatrix}
\cos(\hat{\theta}) & -\sin(\hat{\theta}) & \hat{t}_x \\
\sin(\hat{\theta}) & \cos(\hat{\theta}) & \hat{t}_y
\end{bmatrix}.
\end{align}
This decay-before-drift ordering respects the physical chronology of STEM degradation, where beam-induced damage occurs during exposure, and slow-stage drift affects the entire frame afterward, as shown in Figure \ref{fig:frame}.

\subsection{Training Objective}
\label{sec:degrade}
\begin{figure}[htbp]
    \centering
    \includegraphics[width=1\linewidth]{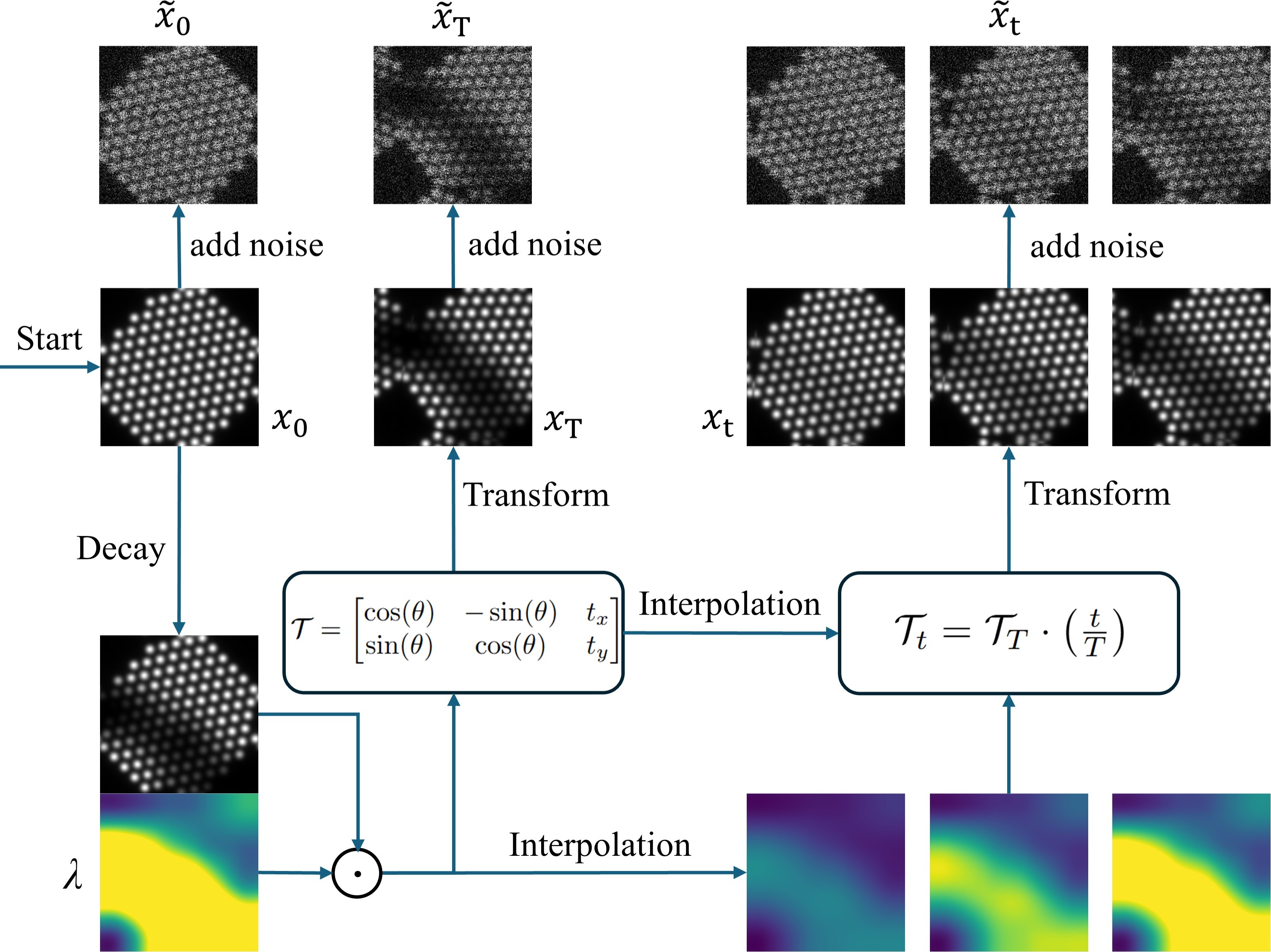}
    \caption{Training process. For each iteration, we start with a random atom map $x_0$, then use the generated decay map $\lambda$ and affine matrix $\mathcal{T}$ to obtain the final state $x_T$. Finally, obtain $x_t$ the interpolation functions.}
    \label{fig:train}
    \vspace{-0.5cm}
\end{figure}

To generate intermediate training samples, we design both the decay map and the drift transformation to be temporally inferred, as shown in Figure~\ref{fig:train}. We start by creating a final decay field $\lambda_T$ using non-linear (Perlin) noise, and compute the intermediate decay as $\lambda_t = \lambda_T \cdot \left(\frac{t}{T}\right)$. The affine drift matrix is similarly defined by interpolating from a final transformation $\mathcal{T}_T$, giving $\mathcal{T}_t = \mathcal{T}_T\cdot \left(\frac{t}{T}\right)$.
This setup allows us to simulate degradation at any time step $t$, with in-batch variation that improves training stability and reduces overfitting.

To better reflect real-world STEM conditions, we add synthetic noise after applying decay and drift, including Poisson noise, scan-line jitter, and readout noise:
\begin{align} 
\tilde{x} = {x} + \epsilon_{\mathrm{noise}}.
\end{align}
Given noisy inputs $(\tilde{x}_0, \tilde{x}_T)$ and a time index $t$, the model is trained to predict the decay map $\hat{\lambda}_t$ and affine transformation $\hat{\mathcal{T}}_t$, which are then used to generate the intermediate prediction $\hat{x}_t$ (see Equation~\ref{eq:pred}).

The training objective is a simple L2 reconstruction loss between the predicted and ground-truth degraded frames:
\begin{align}
\mathcal{L}_{\mathrm{rec}} = \left\| \hat{x}_t - x_t \right\|_2^2.
\end{align}

This setup enables dense supervision and disentangled learning of physical degradation factors, while supporting interpolation or extrapolation across arbitrary time steps $t$. It is well-suited for temporally-aware applications such as future frame prediction or real-time correction in sparse STEM sequences. The overall training pipeline is illustrated in Figure~\ref{fig:train}.

\subsection{Network Architecture and Training Strategy}
The network of AtomDiffuser adopts a dual-stream U-Net backbone, with shared encoders processing the clean input \( x_0 \) and its degraded counterpart \( x_T \) to extract comparable multi-scale features. At the bottleneck, features are fused and modulated by a time embedding \( \gamma(t) \) to enable temporal conditioning. A fully connected head predicts the affine transformation \( \hat{\mathcal{T}}_t \), while the decoder uses dual skip connections to reconstruct a per-pixel decay map \( \hat{\lambda}_t \in [0,1]^{H \times W} \). This architecture integrates spatial structure, temporal context, and global motion into a unified and interpretable framework, with outputs aligned to the input resolution for high-fidelity reconstruction.

Inspired by the data generation strategy in TEMImageNet~\cite{lin2021temimagenet}, we sample raw atom maps (no-background-noise version) from the TEMImageNet dataset (14,364 samples in total) and apply our proposed degradation process dynamically during training to avoid duplication. For each sample, the clean image is used as \( x_0 \), and a degraded version is generated as \( x_T \). The intermediate frame \( x_t \) is synthesized via temporal interpolation, as described in Section~\ref{sec:degrade} and illustrated in Figure~\ref{fig:degrade}.
Our network is implemented in PyTorch and trained on a single NVIDIA RTX 3090 GPU. We use a batch size of 32 and input resolution of \( 1 \times 256 \times 256 \). The model is trained for 200 epochs using the AdamW optimizer with an initial learning rate of 0.0001, decayed progressively with a cosine scheduler to support fine-tuning during the later training stages \cite{zhu2024nnmobilenet,wang2025many}. Training completes in $5.83$ hours under our setup.

\section{Results}
% To evaluate the effectiveness of AtomDiffuser, we conduct quantitative experiments on simulated data with ground-truth supervision for damage prediction, primarily comparing with the existing CNN-based method~\cite{zheng2024convolutional}. We also conduct tests on real cryo-STEM images for quantitative drift estimation. Additionally, we apply our method to real cryo-STEM sequences to produce qualitative results that demonstrate its interpretability and applicability in real-world microscopy scenarios.

\begin{figure*}
    \centering
    \includegraphics[width=1\linewidth]{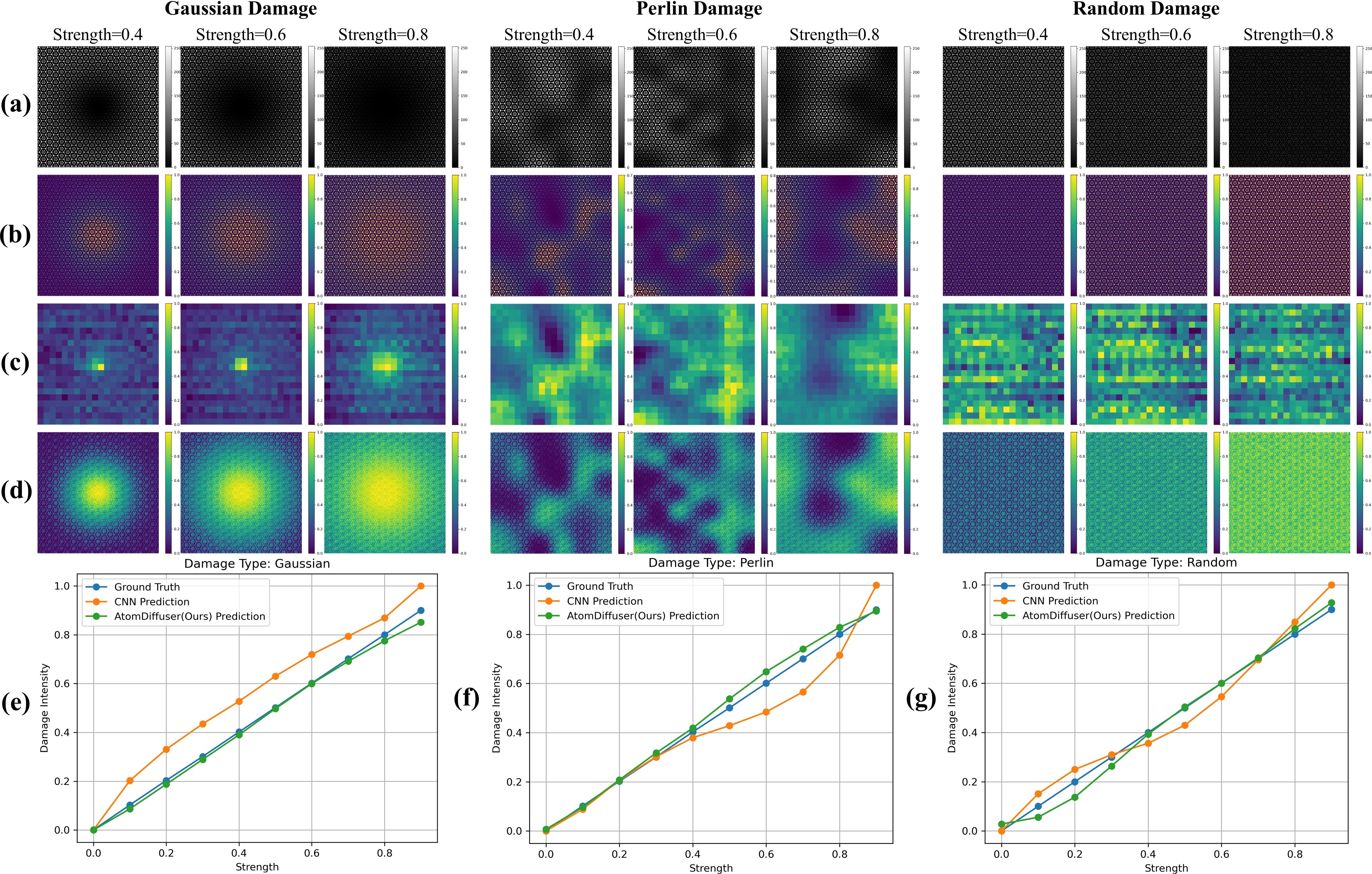}
    \caption{Damage Assessment. (a) are the ground-truth samples, (b) are the ground-truth damage maps, (c) are the predicted damage maps with CNN method \cite{zheng2024convolutional}, (d) are the predicted damage maps with the propsoed method, and (e)(f)(g) are the plotted damage intensity curves with different damage types: Gaussian black hole, Perlin non-linear masks, and Random noise.}
    \label{fig:dmg}
    \vspace{-0.5cm}
\end{figure*}

\subsection{Atomic Damage Assessment}
To evaluate the model’s ability to estimate beam-induced signal degradation independently of spatial drift, we perform a controlled experiment using a fixed synthetic atomic image. We generate degradation-only sequences by applying progressive attenuation while introducing one of three noise types: (1) Gaussian noise with structured black hole artifacts~\cite{zheng2024convolutional}, (2) low-frequency Perlin noise~\cite{wang2024flame}, and (3) random noise mimicking scan jitter and long-range distortions, as illustrated in Figure~\ref{fig:dmg}.
Each sequence consists of ten frames with linearly increasing damage intensity, ranging from $0$ (undamaged) to $0.9$ (severely degraded), and no geometric transformation is applied. This setting isolates the signal decay process, allowing us to assess the model’s capacity to infer spatially varying attenuation under different degradation conditions, as visualized in Figure~\ref{fig:dmg}(e)--(g).

Formally, given a predicted decay map \( \hat{\lambda}_t \), we define the scalar damage intensity as
\begin{equation}
    \bar{\Delta}_t = \frac{1}{HW} \sum_{i=1}^H \sum_{j=1}^W \left(1 - \hat{\lambda}_t(i,j)\right),
\end{equation}
which quantifies the overall proportion of signal loss at time step \( t \). The predicted intensity \( \bar{\Delta}_t \) is then compared against the ground-truth value \( \Delta_t \) across the full temporal sequence.
We compare our model against a CNN-based baseline by plotting the predicted damage intensity curves alongside the ground-truth trajectory, as shown in Figure~\ref{fig:dmg}(e)--(g). Evaluation metrics include mean absolute error (MAE), mean squared error (MSE), root mean squared error (RMSE), and coefficient of determination ($R^2$). Additionally, we report the variance of the prediction error (Var.), computed as the sample variance of the difference between predicted and ground-truth values over time.

To assess the spatial quality of the predicted decay maps, we conduct side-by-side comparisons with a CNN-based baseline~\cite{zheng2024convolutional}. While the CNN performs well under the Perlin setting, our model produces full-resolution predictions aligned with the input image dimensions, enabling fine-grained localization of damage regions, as shown in Figure~\ref{fig:dmg}(d). 
In the Gaussian and Random settings, our method yields more coherent attenuation boundaries and better matches the ground-truth degradation patterns. In contrast, the CNN baseline struggles to generalize under irregular or spatially heterogeneous noise, likely due to resolution loss from downsampling and limited robustness to complex distortions.

\begin{table}[htbp]
\vspace{-0.25cm}
    \centering
    \resizebox{0.8\linewidth}{!}{  \begin{tabular}{c|ccccc}
         &   MAE&MSE&  RMSE&  $R^2$ &Var.\\\toprule
 \multicolumn{6}{c}{Gaussian}\\
 CNN& 0.0994 & 0.0113 & 0.1065& 0.8620
 &0.0177\\
 \textbf{Ours}& \textbf{0.0147}& \textbf{0.0004}& \textbf{0.0198}& \textbf{0.9952} &\textbf{0.0002}\\ \midrule
 \multicolumn{6}{c}{Perlin}\\
         CNN&   0.0573&0.0059&   0.0767&   0.9285 &0.0061\\
         \textbf{Ours}&   \textbf{0.0193}&\textbf{0.0006}&  \textbf{0.0239}&   \textbf{0.9931} &\textbf{0.0003}\\ \midrule
 \multicolumn{6}{c}{Random}\\
 CNN& 0.0635& 0.0074& 0.0859& 0.9106 &0.0025\\
 \textbf{Ours}& \textbf{0.0236}& \textbf{0.0009}& \textbf{0.0306}& \textbf{0.9886} &\textbf{0.0004}\\
    \end{tabular}}
    \caption{Benchmark of Simulated Atomic Damage}
    \label{tab:dmg}
    \vspace{-0.25cm}
\end{table}

\begin{figure*}[htbp]
    \centering
    \includegraphics[width=1\linewidth]{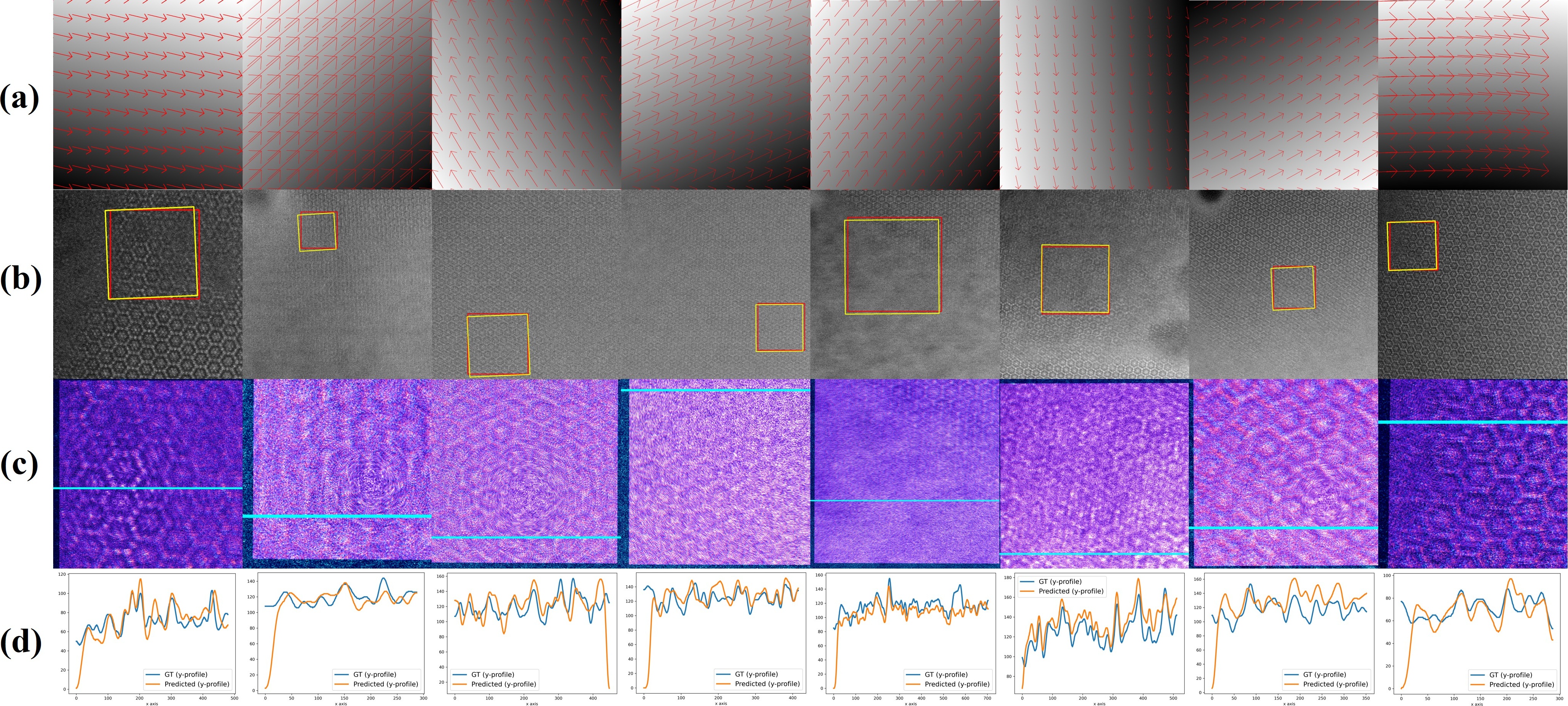}
    \caption{Drift test on STEM samples. (a) are the flow maps of the predicted motion directions, (b) are the STEM samples and its generated two frames ($x_0$ and $x_T$), (c) are the overlaps of predicted $\hat{x}_T$ and $x_T$, where horizontal line indicates the selected location for side-plot, and (d) are the plots of side-views of selected axis from (c).}
    \label{fig:drift}
    \vspace{-0.25cm}
\end{figure*}

To evaluate scalar performance, we test our model on a fixed synthetic sample with known degradation profiles. As shown in Table~\ref{tab:dmg}, our model achieves an MSE as low as $0.0004$ and an $R^2$ of $0.9952$. The predicted intensity closely follows the ground-truth linear trend, demonstrating high accuracy in estimating the degree of beam-induced damage (Figure~\ref{fig:dmg}(e)--(g)).
In contrast, the CNN-based baseline~\cite{zheng2024convolutional}, which treats the task as generic feature-to-feature regression from \( x_0 \) to \( x_t \), exhibits greater variance in cumulative damage estimates. Notably, AtomDiffuser remains robust even under random noise, where the CNN baseline shows significant deviation from the ground-truth trajectory.

\subsection{Sample Drift Correction}
To quantitatively evaluate the model’s ability to recover atomic drift, we conduct a controlled ablation experiment that isolates affine motion without introducing signal decay. Specifically, we disable attenuation (\( \lambda_t = 1 \) everywhere) and test the model’s capacity to predict geometric transformations in real STEM images under synthetic drift conditions.

Given a real high-resolution STEM image \( x_0 \), we randomly extract subregions and apply a simulated affine transformation matrix \( \mathcal{T} \in \mathbb{R}^{2 \times 3} \) to generate a corresponding drifted image \( x_T \):
\begin{equation}
x_T = \Phi(x_0;\ \mathcal{T}),
\end{equation}
where \( \Phi \) denotes an image warping operator parameterized by the ground truth transformation \( \mathcal{T} \). 

For each test, we use our trained model to predict the drift matrix \( \hat{\mathcal{T}} \) that maps \( x_0 \) to \( x_T \). The predicted matrix is then compared against the ground-truth drift matrix \( \mathcal{T} \). We report two evaluation metrics:

\noindent \textbf{Drift error:} the sum of absolute pixel-wise translation errors:
    \begin{equation}
            \mathrm{Drift\ Err.} = |\hat{t}_x - t_x| + |\hat{t}_y - t_y|,
    \end{equation}
    where \( (\hat{t}_x, \hat{t}_y) \) are the translation components of the predicted matrix \( \hat{\mathcal{T}} \), and \( (t_x, t_y) \) are the corresponding components of the ground-truth affine matrix \( \mathcal{T} \).

\noindent \textbf{Rotation error:} the absolute angular difference (in degrees) between the predicted and ground-truth rotation angles:
\begin{equation}
\mathrm{Rotation\ Err.} = |\hat{\theta} - \theta|,
\end{equation}
where \( \hat{\theta} \) and \( \theta \) denote the rotation components of \( \hat{\mathcal{T}} \) and \( \mathcal{T} \), respectively.

Unlike the synthetic degradation setting used during training, this benchmark is constructed entirely from real STEM samples, with single-frame drift derived from geometric analysis.
To visualize the performance, we sampled several test results from each STEM image and plotted detailed comparisons. As shown in Figure~\ref{fig:drift}, the image pairs \( (x_0, x_T) \) are extracted from the original STEM images, and the proposed AtomDiffuser is used to predict the transformation between \( x_0 \) and \( x_T \). The warped version of \( x_T \) is then overlaid onto \( x_0 \) (Figure~\ref{fig:drift}(c)). 

To further assess alignment quality, we visualize a side-profile view (Figure~\ref{fig:drift}(d)). Specifically, we select a random location from the overlapped region in Figure~\ref{fig:drift}(c) and plot the pixel intensity values along a chosen axis. A 1D Gaussian filter is applied to smooth the profiles for improved visualization and interpretability. As shown in Figure~\ref{fig:drift}(d), the drift compensation results in well-aligned intensity peaks between the two signals. In other words, the closer the curves align, the better the model performs in estimating and correcting atomic drift.

\begin{table}[htbp]
    \centering
    \resizebox{1\linewidth}{!}{ \begin{tabular}{cc|cc}\toprule
          Rotation Setting ($^{\circ}$)&Drift Setting (px)&  Drift Err.&  Rotation Err.\\ \toprule
         5&5&  2.4271&  0.6032\\
          10&10&  3.0344&  0.9085\\
          15&15&  3.5103&  1.7986\\
 15& 25& 4.7775& 1.5405\\
 15& 50& 6.4971& 1.3608\\\bottomrule
    \end{tabular}}
    \caption{Drifting Benchmark for STEM Samples}
    \label{tab:drift}
    \vspace{-0.5cm}
\end{table}

As shown in Table~\ref{tab:drift}, the proposed AtomDiffuser successfully estimates affine transformations across a range of synthetic drift and rotation settings applied to real data. The model demonstrates high robustness under moderate motion (e.g., 15~px), achieving low drift errors (2.4--3.5~px) and accurate rotation recovery (below $2^\circ$). 
Importantly, the model maintains stable performance even under asymmetric or unbalanced configurations, such as 15$^\circ$ rotation with 25~px or 50~px drift. In these harder cases, the drift error increases only modestly (to 4.8 and 6.5~px respectively), while rotation accuracy remains largely unaffected. This suggests that the model captures the underlying spatial dynamics without overfitting to balanced motion scales, and generalizes well beyond the training distribution.

\subsection{Degradation Inference \& Performance}
% Although the independent damage assessment and drift estimation perform well in controlled experimental settings, they do not reveal the full potential of the proposed method—the ability to continuously infer between frames and explore material characteristics by observing beam damage progression without the interference of drift.

To demonstrate the full potential of AtomDiffuser—the ability to continuously infer between frames and explore material characteristics by observing beam damage progression without the interference of drift, we apply the model to two consecutive frames selected from a high-resolution cryo-STEM dataset of the garnet-type solid electrolyte \( \mathbf{Li_7La_3Zr_2O_{12}} \) (LLZO)~\cite{zheng2024convolutional}. Specifically, given two frames \( x_0 \) and \( x_T \), the model generates intermediate degradation states, including motion \( \mathcal{T}_t \) and decay factor \( \lambda_t \), by varying the time input \( t \in (0, T) \). Notably, the target frame \( x_T \) exhibits both atomic drift and beam-induced damage relative to \( x_0 \), indicating the two frames are both misaligned and structurally dissimilar.

\begin{figure}[htbp]
    \centering
    \includegraphics[width=1\linewidth]{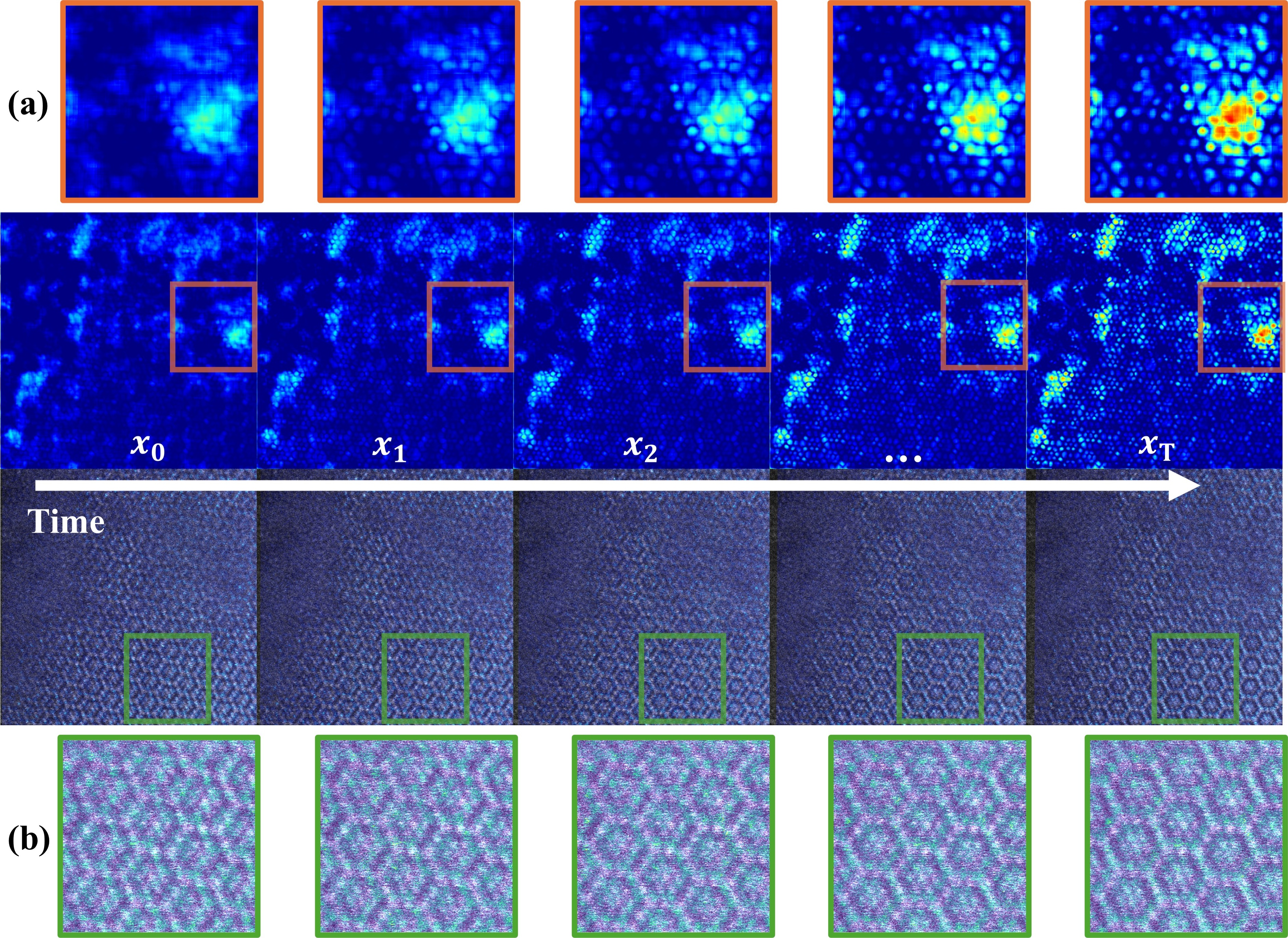}
    \caption{Degradation inference test. For each iteration, we apply $x_0$, $x_T$, and $t$ as inputs, then increase $t$ to generate an $x_t$, subsequently obtained sequences.}
    \label{fig:inter}
    % \vspace{-0.5cm}
\end{figure}

As illustrated in Figure~\ref{fig:inter}(a), the inferred frames exhibit plausible decay progression with smooth transitions in both spatial and intensity domains. Meanwhile, the two frames become gradually aligned as time approaches \( T \), suggesting a recovery of subpixel displacements and damage-aware registration, as shown in Figure~\ref{fig:inter}(b). Overall, the results produced by our method reveal a smooth and interpretable trajectory of structural change that has rarely been demonstrated in existing deep learning approaches.

More importantly, despite the training process being relatively complex and resource-intensive, the model remains lightweight and portable. As shown in Table~\ref{tab:params}, the input size is flexible and supports resolutions up to \( 2048 \times 2048 \) with minimum inference time.
% under our setting (16GB GPU memory), while achieving inference speeds hundreds of times faster than the CNN-based method~\cite{zheng2024convolutional}. 
This makes the model significantly more practical for deployment across various experimental conditions.

\begin{table}[htbp]
    \centering
    \resizebox{1\linewidth}{!}{  \begin{tabular}{c|>{\centering\arraybackslash}p{0.3\linewidth}ccc}\toprule
          &Inference Time (per sample)&  Params (M) & Input Size & Size on disk\\ \toprule
          CNN Method&1.017&  11.69      & 224×224    & 44.59 MB\\
          AtomDiffuser (Ours)&\textbf{0.002}&  \textbf{2.51}& \textbf{2048×2048}    & \textbf{9.98} MB\\\bottomrule
    \end{tabular}}
    \caption{Comparison of Model Parameters}
    \label{tab:params}
    \vspace{-0.6cm}
\end{table}

\section{Conclusion}
Inspired by recent advances in degradation modeling, AtomDiffuser rethinks atomic-scale imaging by modeling degradation as a structured, time-evolving physical process. By disentangling drift and beam-induced damage through a unified, interpretable framework, it enables the observation of structural evolution across sparse STEM sequences. 
Evaluated on both synthetic and cryo-STEM data, AtomDiffuser demonstrates robust performance in recovering atomic motion and damage localization. 
Beyond its performance, this work introduces a new paradigm for degradation modeling in micro-level imaging by leveraging physical priors as conditional signals, which opens new possibilities for studying progressive material evolution in a frame-by-frame technique.
We envision that this capability will support future investigations into time-dependent beam effects, enable virtual time-lapse reconstruction from sparse acquisitions, and provide new insights into dynamic material behavior at the atomic scale.

\section*{Acknowledgments}
The authors acknowledge the support by the National Science Foundation under Award No. 2239598, as well as the use of facilities and instrumentation at the University of California Irvine Materials Research Institute, which is supported in part by the National Science Foundation Materials Research Science and Engineering Center program through the University of California Irvine Center for Complex and Active Materials (DMR-2011967).

{
    \small
    \bibliographystyle{ieeenat_fullname}
    \bibliography{main}
}


\end{document}

%% file: main.bbl
\begin{thebibliography}{40}
\providecommand{\natexlab}[1]{#1}
\providecommand{\url}[1]{\texttt{#1}}
\expandafter\ifx\csname urlstyle\endcsname\relax
  \providecommand{\doi}[1]{doi: #1}\else
  \providecommand{\doi}{doi: \begingroup \urlstyle{rm}\Url}\fi

\bibitem[Bansal et~al.(2023)Bansal, Borgnia, Chu, Li, Kazemi, Huang, Goldblum, Geiping, and Goldstein]{bansal2023cold}
Arpit Bansal, Eitan Borgnia, Hong-Min Chu, Jie Li, Hamid Kazemi, Furong Huang, Micah Goldblum, Jonas Geiping, and Tom Goldstein.
\newblock Cold diffusion: Inverting arbitrary image transforms without noise.
\newblock \emph{Advances in Neural Information Processing Systems}, 36:\penalty0 41259--41282, 2023.

\bibitem[Belardi et~al.(2025)Belardi, Lee, Wang, Lovelace, Weinberger, Muller, and Gomes]{belardi2025improving}
Christian~K Belardi, Chia-Hao Lee, Yingheng Wang, Justin Lovelace, Kilian~Q Weinberger, David~A Muller, and Carla~P Gomes.
\newblock Improving multislice electron ptychography with a generative prior.
\newblock \emph{arXiv preprint arXiv:2507.17800}, 2025.

\bibitem[Choi et~al.(2023)Choi, Choi, Kim, Kim, and Yoon]{choi2023custom}
Jooyoung Choi, Yunjey Choi, Yunji Kim, Junho Kim, and Sungroh Yoon.
\newblock Custom-edit: Text-guided image editing with customized diffusion models.
\newblock \emph{arXiv preprint arXiv:2305.15779}, 2023.

\bibitem[Dey et~al.(2024)Dey, De~Ridder, Blanco, Halder, and Van~Waeyenberge]{dey2024addressing}
Bappaditya Dey, Vic De~Ridder, Victor Blanco, Sandip Halder, and Bartel Van~Waeyenberge.
\newblock Addressing class imbalance and data limitations in advanced node semiconductor defect inspection: A generative approach for sem images.
\newblock In \emph{2024 International Symposium ELMAR}, pages 141--148. IEEE, 2024.

\bibitem[Egerton et~al.(2004)Egerton, Li, and Malac]{egerton2004radiation}
RF Egerton, P Li, and M Malac.
\newblock Radiation damage in the tem and sem.
\newblock \emph{Micron}, 35\penalty0 (6):\penalty0 399--409, 2004.

\bibitem[Ghosh et~al.(2022)Ghosh, Ziatdinov, Dyck, Sumpter, and Kalinin]{ghosh2022bridging}
Ayana Ghosh, Maxim Ziatdinov, Ondrej Dyck, Bobby~G Sumpter, and Sergei~V Kalinin.
\newblock Bridging microscopy with molecular dynamics and quantum simulations: an atomai based pipeline.
\newblock \emph{npj Computational Materials}, 8\penalty0 (1):\penalty0 74, 2022.

\bibitem[He et~al.(2021)He, Zhang, Zhang, and Han]{he2021hybrid}
Bintao He, Fa Zhang, Huanshui Zhang, and Renmin Han.
\newblock A hybrid frequency-spatial domain model for sparse image reconstruction in scanning transmission electron microscopy.
\newblock In \emph{Proceedings of the IEEE/CVF International Conference on Computer Vision}, pages 2682--2691, 2021.

\bibitem[Hussaini et~al.(2018)Hussaini, Lin, Natarajan, Zhu, and Sharma]{hussaini2018determination}
Zahra Hussaini, Pin~Ann Lin, Bharath Natarajan, Wenhui Zhu, and Renu Sharma.
\newblock Determination of atomic positions from time resolved high resolution transmission electron microscopy images.
\newblock \emph{Ultramicroscopy}, 186:\penalty0 139--145, 2018.

\bibitem[Ihara et~al.(2022)Ihara, Saito, Yoshinaga, Avala, and Murayama]{ihara2022deep}
Shiro Ihara, Hikaru Saito, Mizumo Yoshinaga, Lavakumar Avala, and Mitsuhiro Murayama.
\newblock Deep learning-based noise filtering toward millisecond order imaging by using scanning transmission electron microscopy.
\newblock \emph{Scientific reports}, 12\penalty0 (1):\penalty0 13462, 2022.

\bibitem[Kalinin et~al.(2021)Kalinin, Ziatdinov, Hinkle, Jesse, Ghosh, Kelley, Lupini, Sumpter, and Vasudevan]{kalinin2021automated}
Sergei~V Kalinin, Maxim Ziatdinov, Jacob Hinkle, Stephen Jesse, Ayana Ghosh, Kyle~P Kelley, Andrew~R Lupini, Bobby~G Sumpter, and Rama~K Vasudevan.
\newblock Automated and autonomous experiments in electron and scanning probe microscopy.
\newblock \emph{ACS nano}, 15\penalty0 (8):\penalty0 12604--12627, 2021.

\bibitem[Lee et~al.(2025)Lee, Khan, Clark, and Huang]{lee2025using}
Chia-Hao Lee, Abid Khan, Bryan Clark, and Pinshane~Y Huang.
\newblock Using cyclegans to generate realistic stem images for machine learning and atom-by-atom analysis on the million-atom scale.
\newblock \emph{Microscopy and Microanalysis}, 31\penalty0 (Supplement\_1):\penalty0 ozaf048--1074, 2025.

\bibitem[Liao et~al.(2024)Liao, Xu, Liu, Gao, Jin, Liang, and Lu]{liao2024fast}
Ruihao Liao, Ke Xu, Yifan Liu, Zibo Gao, Shuo Jin, Linyun Liang, and Guang-Hong Lu.
\newblock Fast prediction of irradiation-induced cascade defects using denoising diffusion probabilistic model.
\newblock \emph{Nuclear Materials and Energy}, 41:\penalty0 101805, 2024.

\bibitem[Lin et~al.(2021)Lin, Zhang, Wang, Yang, and Xin]{lin2021temimagenet}
Ruoqian Lin, Rui Zhang, Chunyang Wang, Xiao-Qing Yang, and Huolin~L Xin.
\newblock Temimagenet training library and atomsegnet deep-learning models for high-precision atom segmentation, localization, denoising, and deblurring of atomic-resolution images.
\newblock \emph{Scientific reports}, 11\penalty0 (1):\penalty0 5386, 2021.

\bibitem[Lobato et~al.(2024)Lobato, Friedrich, and Van~Aert]{lobato2024deep}
I Lobato, T Friedrich, and S Van~Aert.
\newblock Deep convolutional neural networks to restore single-shot electron microscopy images.
\newblock \emph{npj Computational Materials}, 10\penalty0 (1):\penalty0 10, 2024.

\bibitem[Lu et~al.(2023)Lu, Chen, Qiu, Chen, Chen, Qi, and Jiang]{lu2023emdiffuse}
Chixiang Lu, Kai Chen, Heng Qiu, Xiaojun Chen, Gu Chen, Xiaojuan Qi, and Haibo Jiang.
\newblock Emdiffuse: a diffusion-based deep learning method augmenting ultrastructural imaging and volume electron microscopy.
\newblock \emph{bioRxiv}, pages 2023--07, 2023.

\bibitem[Lu et~al.(2024)Lu, Chen, Qiu, Chen, Chen, Qi, and Jiang]{lu2024diffusion}
Chixiang Lu, Kai Chen, Heng Qiu, Xiaojun Chen, Gu Chen, Xiaojuan Qi, and Haibo Jiang.
\newblock Diffusion-based deep learning method for augmenting ultrastructural imaging and volume electron microscopy.
\newblock \emph{Nature Communications}, 15\penalty0 (1):\penalty0 4677, 2024.

\bibitem[Masud et~al.(2024)Masud, Rade, Hasib, Krishnamurthy, and Sarkar]{masud2024machine}
Nabila Masud, Jaydeep Rade, Md~Hasibul~Hasan Hasib, Adarsh Krishnamurthy, and Anwesha Sarkar.
\newblock Machine learning approaches for improving atomic force microscopy instrumentation and data analytics.
\newblock \emph{Frontiers in Physics}, 12:\penalty0 1347648, 2024.

\bibitem[Meng et~al.(2021)Meng, He, Song, Song, Wu, Zhu, and Ermon]{meng2021sdedit}
Chenlin Meng, Yutong He, Yang Song, Jiaming Song, Jiajun Wu, Jun-Yan Zhu, and Stefano Ermon.
\newblock Sdedit: Guided image synthesis and editing with stochastic differential equations.
\newblock \emph{arXiv preprint arXiv:2108.01073}, 2021.

\bibitem[Mokady et~al.(2023)Mokady, Hertz, Aberman, Pritch, and Cohen-Or]{mokady2023null}
Ron Mokady, Amir Hertz, Kfir Aberman, Yael Pritch, and Daniel Cohen-Or.
\newblock Null-text inversion for editing real images using guided diffusion models.
\newblock In \emph{Proceedings of the IEEE/CVF Conference on Computer Vision and Pattern Recognition}, pages 6038--6047, 2023.

\bibitem[Ophus et~al.(2016)Ophus, Ciston, and Nelson]{ophus2016correcting}
Colin Ophus, Jim Ciston, and Chris~T Nelson.
\newblock Correcting nonlinear drift distortion of scanning probe and scanning transmission electron microscopies from image pairs with orthogonal scan directions.
\newblock \emph{Ultramicroscopy}, 162:\penalty0 1--9, 2016.

\bibitem[Pakornchote et~al.(2024)Pakornchote, Choomphon-Anomakhun, Arrerut, Atthapak, Khamkaeo, Chotibut, and Bovornratanaraks]{pakornchote2024diffusion}
Teerachote Pakornchote, Natthaphon Choomphon-Anomakhun, Sorrjit Arrerut, Chayanon Atthapak, Sakarn Khamkaeo, Thiparat Chotibut, and Thiti Bovornratanaraks.
\newblock Diffusion probabilistic models enhance variational autoencoder for crystal structure generative modeling.
\newblock \emph{Scientific Reports}, 14\penalty0 (1):\penalty0 1275, 2024.

\bibitem[Qiu et~al.(2025)Qiu, Meng, Li, Hong, Li, Han, Liang, Cheng, Ke, Zhang, et~al.]{qiu2025deep}
Zanlin Qiu, Yuan Meng, Junxian Li, Yanhui Hong, Ning Li, Xiaocang Han, Yu Liang, Wing~Ni Cheng, Guolin Ke, Linfeng Zhang, et~al.
\newblock Deep learning for sub-{\aa}ngstr{\"o}m resolution imaging in uncorrected scanning transmission electron microscope.
\newblock \emph{National Science Review}, page nwaf235, 2025.

\bibitem[Roccapriore et~al.(2021)Roccapriore, Creange, Ziatdinov, and Kalinin]{roccapriore2021identification}
Kevin~M Roccapriore, Nicole Creange, Maxim Ziatdinov, and Sergei~V Kalinin.
\newblock Identification and correction of temporal and spatial distortions in scanning transmission electron microscopy.
\newblock \emph{Ultramicroscopy}, 229:\penalty0 113337, 2021.

\bibitem[Rombach et~al.(2022)Rombach, Blattmann, Lorenz, Esser, and Ommer]{rombach2022high}
Robin Rombach, Andreas Blattmann, Dominik Lorenz, Patrick Esser, and Bj{\"o}rn Ommer.
\newblock High-resolution image synthesis with latent diffusion models.
\newblock In \emph{Proceedings of the IEEE/CVF conference on computer vision and pattern recognition}, pages 10684--10695, 2022.

\bibitem[Sadri et~al.(2024)Sadri, Petersen, Terzoudis-Lumsden, Esser, Etheridge, and Findlay]{sadri2024unsupervised}
Alireza Sadri, Timothy~C Petersen, Emmanuel~WC Terzoudis-Lumsden, Bryan~D Esser, Joanne Etheridge, and Scott~D Findlay.
\newblock Unsupervised deep denoising for four-dimensional scanning transmission electron microscopy.
\newblock \emph{npj Computational Materials}, 10\penalty0 (1):\penalty0 243, 2024.

\bibitem[Sang and LeBeau(2014)]{sang2014revolving}
Xiahan Sang and James~M LeBeau.
\newblock Revolving scanning transmission electron microscopy: Correcting sample drift distortion without prior knowledge.
\newblock \emph{Ultramicroscopy}, 138:\penalty0 28--35, 2014.

\bibitem[Sang et~al.(2017)Sang, Lupini, Ding, Kalinin, Jesse, and Unocic]{sang2017precision}
Xiahan Sang, Andrew~R Lupini, Jilai Ding, Sergei~V Kalinin, Stephen Jesse, and Raymond~R Unocic.
\newblock Precision controlled atomic resolution scanning transmission electron microscopy using spiral scan pathways.
\newblock \emph{Scientific Reports}, 7\penalty0 (1):\penalty0 43585, 2017.

\bibitem[Wang et~al.(2024)Wang, Boroujeni, Chen, Bastola, Li, Zhu, and Razi]{wang2024flame}
Hao Wang, Sayed Pedram~Haeri Boroujeni, Xiwen Chen, Ashish Bastola, Huayu Li, Wenhui Zhu, and Abolfazl Razi.
\newblock Flame diffuser: Wildfire image synthesis using mask guided diffusion.
\newblock In \emph{2024 IEEE International Conference on Big Data (BigData)}, pages 6171--6179. IEEE, 2024.

\bibitem[Wang et~al.(2025{\natexlab{a}})Wang, Chen, Bastola, Qin, and Razi]{Wang_2025_WACV}
Hao Wang, Xiwen Chen, Ashish Bastola, Jiayou Qin, and Abolfazl Razi.
\newblock Diffusion prism: Enhancing diversity and morphology consistency in mask-to-image diffusion.
\newblock In \emph{Proceedings of the Winter Conference on Applications of Computer Vision (WACV) Workshops}, pages 228--237, 2025{\natexlab{a}}.

\bibitem[Wang et~al.(2025{\natexlab{b}})Wang, Zhu, Dong, Chen, Li, Qiu, Chen, Vasa, Xiong, Dumitrascu, et~al.]{wang2025many}
Hao Wang, Wenhui Zhu, Xuanzhao Dong, Yanxi Chen, Xin Li, Peijie Qiu, Xiwen Chen, Vamsi~Krishna Vasa, Yujian Xiong, Oana~M Dumitrascu, et~al.
\newblock Many-mobilenet: Multi-model augmentation for robust retinal disease classification.
\newblock In \emph{MICCAI Challenge on Ultra-Widefield Fundus Imaging for Diabetic Retinopathy}, pages 144--154. Springer, 2025{\natexlab{b}}.

\bibitem[Weile et~al.(2025)Weile, Grytsiuk, Penn, Chica, Roy, Mosina, Sofer, Schi{\o}tz, Helveg, R{\"o}sner, et~al.]{weile2025defect}
Mads Weile, Sergii Grytsiuk, Aubrey Penn, Daniel~G Chica, Xavier Roy, Kseniia Mosina, Zdenek Sofer, Jakob Schi{\o}tz, Stig Helveg, Malte R{\"o}sner, et~al.
\newblock Defect complexes in crsbr revealed through electron microscopy and deep learning.
\newblock \emph{Physical Review X}, 15\penalty0 (2):\penalty0 021080, 2025.

\bibitem[Xiao et~al.(2024)Xiao, Zhang, and Li]{xiao2024automated}
Rui Xiao, Yanzhu Zhang, and Mi Li.
\newblock Automated high-throughput atomic force microscopy single-cell nanomechanical assay enabled by deep learning-based optical image recognition.
\newblock \emph{Nano Letters}, 24\penalty0 (39):\penalty0 12323--12332, 2024.

\bibitem[Ye et~al.(2023)Ye, Wang, Zhang, Gao, Wang, and Sun]{ye2023recovering}
Enze Ye, Yuhang Wang, Hong Zhang, Yiqin Gao, Huan Wang, and He Sun.
\newblock Recovering a molecule's 3d dynamics from liquid-phase electron microscopy movies.
\newblock In \emph{Proceedings of the IEEE/CVF International Conference on Computer Vision}, pages 10767--10777, 2023.

\bibitem[Zhang et~al.(2025)Zhang, Wen, Han, Juefei-Xu, Srivastava, Huang, Pavlovic, Wang, Tao, and Metaxas]{zhang2025soda}
Xinxi Zhang, Song Wen, Ligong Han, Felix Juefei-Xu, Akash Srivastava, Junzhou Huang, Vladimir Pavlovic, Hao Wang, Molei Tao, and Dimitris Metaxas.
\newblock Soda: Spectral orthogonal decomposition adaptation for diffusion models.
\newblock In \emph{2025 IEEE/CVF Winter Conference on Applications of Computer Vision (WACV)}, pages 4665--4682. IEEE, 2025.

\bibitem[Zheng et~al.(2022)Zheng, Lu, and He]{zheng2022situ}
Hongkui Zheng, Xiner Lu, and Kai He.
\newblock In situ transmission electron microscopy and artificial intelligence enabled data analytics for energy materials.
\newblock \emph{Journal of Energy Chemistry}, 68:\penalty0 454--493, 2022.

\bibitem[Zheng et~al.(2024)Zheng, Chen, Razi, and He]{zheng2024convolutional}
Hongkui Zheng, Xiwen Chen, Abolfazl Razi, and Kai He.
\newblock Convolutional neural networks for evaluation of sequential beam damage of beam-sensitive solid electrolytes, 2024.

\bibitem[Zheng et~al.(2025)Zheng, Wang, Chen, Razi, and He]{zheng2025gaussian}
Hongkui Zheng, Hao Wang, Xiwen Chen, Abolfazl Razi, and Kai He.
\newblock Gaussian differential assessment of sequential stem radiation damage in beam-sensitive materials.
\newblock \emph{Microscopy and Microanalysis}, 31\penalty0 (Supplement\_1):\penalty0 ozaf048--1065, 2025.

\bibitem[Zhou et~al.(2023)Zhou, Lou, Khanna, and Ermon]{zhou2023denoising}
Linqi Zhou, Aaron Lou, Samar Khanna, and Stefano Ermon.
\newblock Denoising diffusion bridge models.
\newblock \emph{arXiv preprint arXiv:2309.16948}, 2023.

\bibitem[Zhu et~al.(2024)Zhu, Qiu, Chen, Li, Lepore, Dumitrascu, and Wang]{zhu2024nnmobilenet}
Wenhui Zhu, Peijie Qiu, Xiwen Chen, Xin Li, Natasha Lepore, Oana~M Dumitrascu, and Yalin Wang.
\newblock nnmobilenet: Rethinking cnn for retinopathy research.
\newblock In \emph{Proceedings of the IEEE/CVF Conference on Computer Vision and Pattern Recognition}, pages 2285--2294, 2024.

\bibitem[Ziatdinov et~al.(2022)Ziatdinov, Ghosh, Wong, and Kalinin]{ziatdinov2022atomai}
Maxim Ziatdinov, Ayana Ghosh, Chun~Yin Wong, and Sergei~V Kalinin.
\newblock Atomai framework for deep learning analysis of image and spectroscopy data in electron and scanning probe microscopy.
\newblock \emph{Nature Machine Intelligence}, 4\penalty0 (12):\penalty0 1101--1112, 2022.

\end{thebibliography}
